\documentclass{article}

\usepackage{arxiv}

\usepackage[utf8]{inputenc} % allow utf-8 input
\usepackage[T1]{fontenc}    % use 8-bit T1 fonts
\usepackage{url}            % simple URL typesetting
\usepackage{booktabs}       % professional-quality tables
\usepackage{amsfonts}       % blackboard math symbols
\usepackage{nicefrac}       % compact symbols for 1/2, etc.
\usepackage{microtype}      % microtypography
\usepackage{lipsum}
\usepackage{graphicx}

\usepackage{amssymb} % Package for math lettering
\usepackage{acronym}
\usepackage{amsmath}
\usepackage{booktabs} % For pretty tables
\usepackage{caption} % For caption spacing
\usepackage{subcaption} % For sub-figures
\usepackage{pgfplots}
\acrodef{NAS}[NAS]{Neuromorphic Auditory Sensor}
\acrodef{SSP}[SSP]{spike signals processing}
\acrodef{FPGA}[FPGA]{Field-Programmable Gate Array}
\acrodef{AER}[AER]{Address-Event Representation}
\acrodef{SSG}[SSG]{synthetic spike generator}

\title{Event Based Time-Vectors for auditory features extraction:
        a neuromorphic approach for low power audio recognition}

\author{
 Marco Rasetto \\
  Department of Bioengineering and Center for Neural Basis of Cognition\\
  University of Pittsburgh, Carnegie Mellon University\\
  Pittsburgh, USA \\
  \texttt{marcorasetto@pitt.edu} \\
   \And
 Juan P. Dominguez-Morales \\
  Robotics and Technology of Computers Lab \\
  Universidad de Sevilla\\
  Seville, Spain \\
  \And
 Angel Jimenez-Fernandez \\
  Robotics and Technology of Computers Lab \\
  Universidad de Sevilla\\
  Seville, Spain \\
  \And
 Ryad Benosman \\
  Department of Bioengineering\\
  University of Pittsburgh\\
  Pittsburgh, USA \\
  \texttt{benosman@pitt.edu}
}

\begin{document}
\maketitle
\begin{abstract}
In recent years tremendous efforts have been done to advance the state of the art for Natural Language Processing (NLP) and audio recognition. However, these efforts often translated in increased power consumption and memory requirements for bigger and more complex models. These solutions falls short of the constraints of IoT devices which need low power, low memory efficient computation, and therefore they fail to meet the growing demand of efficient edge computing. Neuromorphic systems have proved to be excellent candidates for low-power low-latency computation in a multitude of applications. For this reason we present a neuromorphic architecture, capable of unsupervised auditory feature recognition.
We then validate the network on a subset of Google's Speech Commands dataset.
\end{abstract}

% keywords can be removed


\section{Introduction}

%A first strategy to work with a NAS is to adapt $\tau_{ch_i}$ to accommodate the natural difference of gain among the channels of the cochlea. Similarly to a natural cochlea, the NAS presents a dampening effect caused by the cascade of filters, with lower frequency filters producing less events. Given the progressive decrease of spiking rate from high frequency to low frequency channels, the local time surfaces need to be adjusted accordingly.%

With ever growing markets of the Internet-Of-Things \cite{edgeAI} \cite{IOT}, portable and implantable medical devices\cite{market2018global}\cite{medicalkumari2017increasing}, and mobile robotics\cite{harrop2016mobile}, there is an increasing need of low power AI algorithms and systems able to operate with low latency and low memory requirements. One perfect example of this trend is the field of natural language processing for virtual assistants, where end users interact with generally low power devices that, after recognizing a wakeword (''Hey google''), require constant connection to servers in order to respond to commands and run more complex machine learning models \cite{chen2019deep}.
We argue that a solution to this problem of efficiency could be found in Neuromorphic systems.
Neuromorphic engineering is a concept coined by Carver Mead in the late eighties \cite{mead1990neuromorphic}, it refers to the design and implementation of hardware and software to mimic the basic principles of biological nervous systems. In this approach, information is represented and transmitted by means of asynchronous pulses or action potentials emitted by artificial neurons separated in building blocks or different event-based processing layers. This architecture presents a clear advantage over traditional computation methods in terms of power consumption and real-time capabilities \cite{indiveri2011neuromorphic}\cite{NeuromorphNature}\cite{zhu2020comprehensiveReview}. %Previous works have presented implementations of novel neuromorphic systems for robotic applications, in which the two aforementioned advantages are of key relevance. Some of them are focused on ... 

%With the field of neuromorphic computation steadily growing, the advent of event-based cameras posed a new incentive in rethinking the algorithms used in computer visions that were often insensitive to temporal information of the stimuli and based on a synchronous framed decomposition of the visual information. This approach helped to develop algorithms that where faster, with lower latency and more scalable than standard counterparts [Citation needed here, maybe Gregor blinking algorithm and HOTS paper].

In order to fully appreciate the power efficiency of these systems, however, it is necessary to design their architecture from the ground up, starting from sensors. In the case of neuromorphic auditory systems, the sensor would be a bio-inspired cochlea.
Bio-inspired cochleas are sensors which are inspired in how the biological human ear behaves, decomposing the input audio signal into different streams of events representing different frequency bands. Different software and hardware implementations of bio-inspired cochleas can be found in the literature, with all of them having a similar architecture: first, an input stage is used to receive the stimuli; then, a set of band-pass filters (following either a cascade or a parallel topology) is used to process the information; and, finally, these streams are outputted. Some examples of state-of-the-art event-based cochleas are \cite{yang20160,chan2007aer} (analog) and  \cite{xu2018fpga,jimenez2016binaural} (digital). Additional processing steps can be applied to the output of these sensors for further natural language processing, sound recognition \cite{o2013real} and sound source localization tasks 
\cite{schoepe2019neuromorphic}, among others.
In this paper we propose a novel neuromorphic architecture for sound recognition feature extraction, designed to be compatible with a digital cochlea designed by Jimenez et al \cite{jimenez2016binaural}. The proposed network is designed to work on a per event basis, learning patterns of relative timings of activation generated by the cochlea in a unsupervised manner. 
When complemented with a classifier, the network can be used for phoneme/word recognition, making it an important first stage for designing full Neuromorphic systems for low power, low latency NLP.
Whereas similar solutions were proposed on visual neuromorphic architectures \cite{lagorce2016hots} \cite{hots2018sparse}, to our knowledge, this is the first time that this approach has ever been used for auditory pattern recognition.

%[Talk about state-of-the-art feature extraction in event-based systems]
%Regarding state-of-the-art, some papers that we should cite are: \cite{linares2019low,afshar2019investigation,afshar2020event,pan2018event}.

%The rest of the paper is structured as follows: [... write something here depending on the structure of the final paper.]

\section{Materials and Methods}

\subsection{Dataset}
In this work, Google's Speech Commands Dataset \cite{warden2018speech} was used. This dataset consists of 65000 one-second long utterances of 30 short words, by thousands of different people. In our case, the "on" (3846 samples) and "off" (3746 samples) classes were used to train and test the proposed system. 

Each sample was fed to a 32-channel monaural \ac{NAS} and then saved as an AEDAT file (consisting of addresses and timestamps of the spikes produced) using an USBAERmini2 board. Therefore, a total of 7592 AEDAT files with the spiking information decomposed in 32 different frequency bands were obtained.

\subsection{Neuromorphic Auditory Sensor}
A \ac{NAS} \cite{jimenez2016binaural} is a digital cochlea implementation that is based in Lyon’s cascaded model \cite{lyon1982computational} of the biological cochlea and also on \ac{SSP} techniques \cite{jimenez2010building,jimenez2012neuro}. \ac{NAS} decomposes the input audio obtained from an analog-to-digital audio converter with I2S interface in its frequency components, and outputs a stream of events using the \ac{AER}\cite{AER}.

After the input sound signal is digitized using the audio converter, the information is written sample-by-sample in a \ac{SSG} \cite{paz2009synthetic}, which provides a spike stream with a frequency that is proportional to the amplitude of the input audio signal. Then, the spike stream passes through a cascaded bank of spike-based band-pass filters that perform the frequency decomposition. The output of each of the filters is connected to an \ac{AER}-monitor, which assigns a unique address to the spikes following the \ac{AER} representation. It is important to mention that for every filter (also called channel), two different unique addresses are generated: one of them for the spikes corresponding to the positive part of the signal and the other one for the negative part. Finally, the spikes are propagated through an asynchronous \ac{AER} bus to the output, being able to interface with other neuromorphic platforms such as SpiNNaker\cite{furber2014spinnaker}, \acp{FPGA} or other \ac{AER}-based boards.

\begin{figure}[ht]
    \centering
    \includegraphics[scale=0.7]{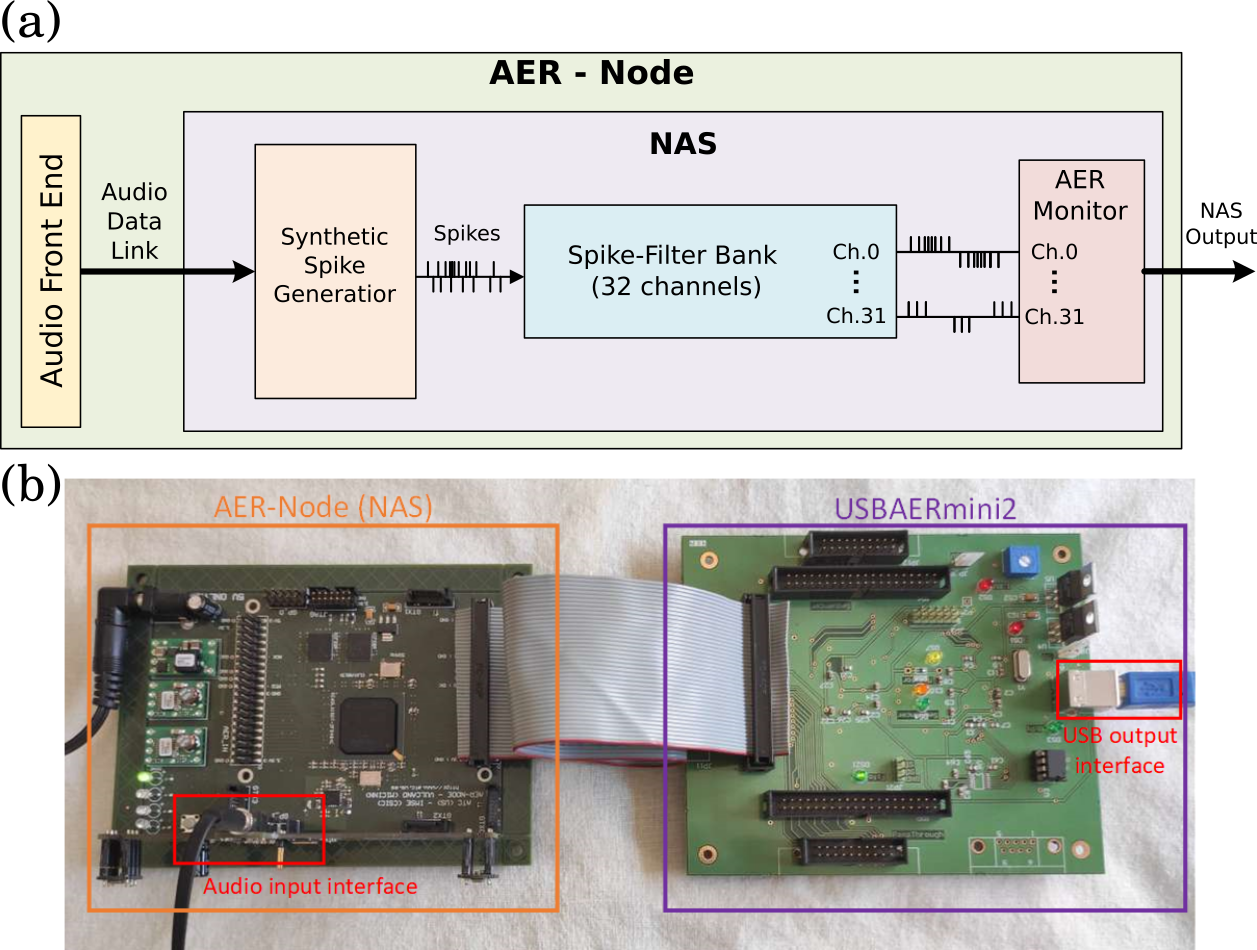}
    \caption{(a) Block diagram of the NAS implementation on an AER-Node board.(b) A picture of the AER-Node board, and a USBAERmini2, used to sample and save the NAS Output on a PC}
    \label{fig:NAS_block_diagram}
\end{figure}

In this work, a 32-channels (64 different addresses) monaural \ac{NAS} was implemented in an AER-Node board, which has a Spartan-6 \ac{FPGA}. A block diagram of the NAS implementation can be seen in Fig. \ref{fig:NAS_block_diagram}. The 32 channels of the implemented \ac{NAS} are distributed along the whole range of the human hearing (20 Hz to 20 kHz, approximately \cite{rosen2011signals}).  The output of the \ac{NAS} is connected to an USBAERmini2 board \cite{berner20075} that assigns a timestamp to the spikes and allows to log the spiking information on the computer by using either jAER\footnote{http://jaerproject.org (accessed \today)} or Matlab as AER-DATA (.aedat) files.
As the negative addresses are considered redundant for the scenario of auditory data classification, they are discarded before being relayed to the proposed Neuromorphic Network.

\subsection{Network Architecture}
\begin{figure}[ht]
    \centering
    \includegraphics[scale=0.66]{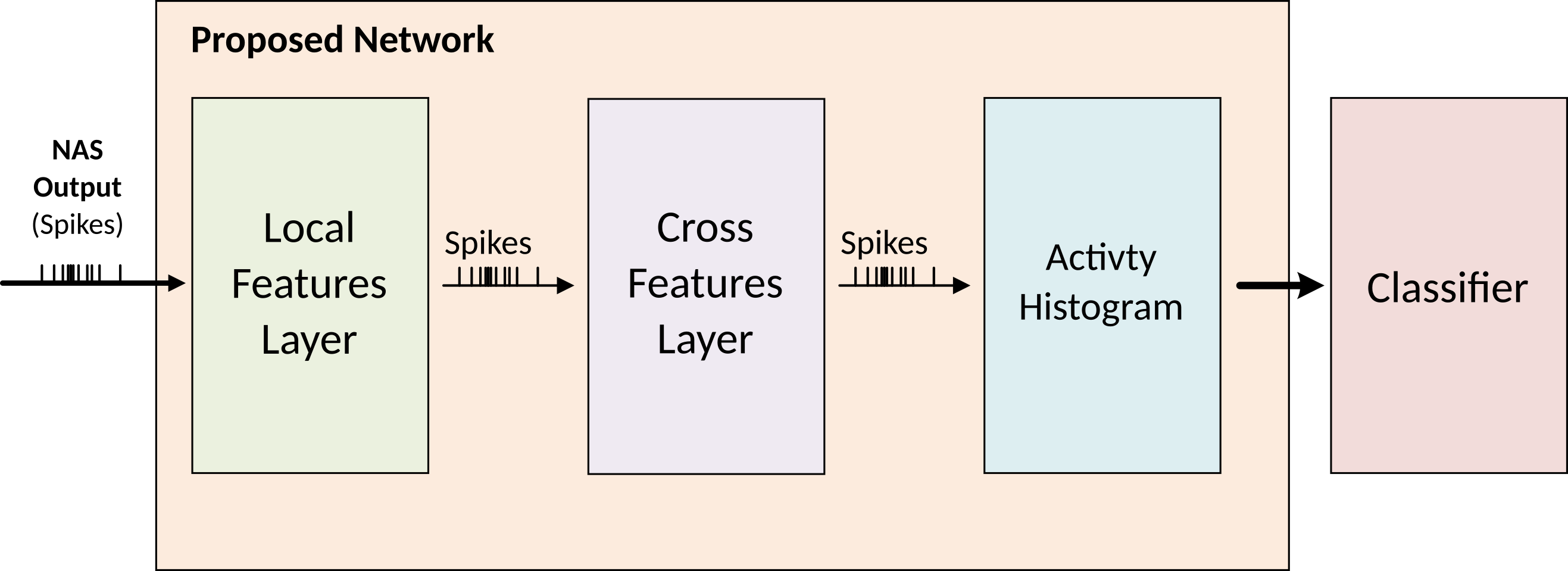}
    \caption{Block diagram of the proposed event based network. NAS output Spikes are fed to the Local Features Layer extracting temporal patterns of activity of single channels. Its output is then fed to the Cross Feature Layer that learns to extract patterns of activity across different channels. The Output of the Cross Feature Layers is then used to generate histograms of activity that can finally be classified in order to recognize the input stimuli.}
    \label{fig:Gordonn}
\end{figure}

Neuromorphic cochleas, like biological counterparts, decompose auditory stimuli in a series of fundamental frequencies and encode the sound pressure with spikes. In the particular case of the NAS, sound pressure is first encoded as rate, transformed in spiking data, and then filtered in different channels using Spike-Filter Banks as shown in Figure \ref{fig:NAS_block_diagram}. This process causes the spiking rate of higher frequency channels to be proportionally higher than lower frequency channels \cite{jimenez2016binaural}. This difference in spiking rate is hard to treat when designing neuromorphic algorithms that deal with precise timings of activation, as they often expect the same integration time scale to be shared among neuron populations or units of the network.
Our network (Fig. \ref{fig:Gordonn}) tackles this problem by allowing spiking information to be integrated at two separate stages: a Local Features Layer and a Cross Features Layer. In the Local layer we extract pattern of activity of the NAS of individual channels, and we build some descriptors called Local Time-Vectors that will be normalized to deal with the spiking rate difference. These descriptors are what is used by the Local Features Layer to learn and recognize spiking patterns from single channels. The output of the Local Features Layer is then pooled across channels using another descriptor called Cross Time-Vector that is then fed to the Cross Feature Layer, with the purpose of classifying patterns of the whole Neuromorphic Cochlea.
The Cross Features Layer output is then used to build Histograms of Activity that can be then used as input for a classifier with the purpose of recognizing the input stimuli to the whole network. 
In the following paragraphs we will explain the details of Time-Vector generation and how individual layers of the network work.

\subsection{Local features layer}
In this section we define the first layer of the architecture, aimed to extract temporal patterns of single channels of the NAS. We first define the concept of Local Time-Vector, and then we proceed to explain how we can learn recurring patterns and use this descriptor to generate a novel stream of events that will be processed by the second layer.

\subsubsection{Local Time-Vectors:}
\begin{figure}[ht]
\centering
    \includegraphics[scale=0.62]{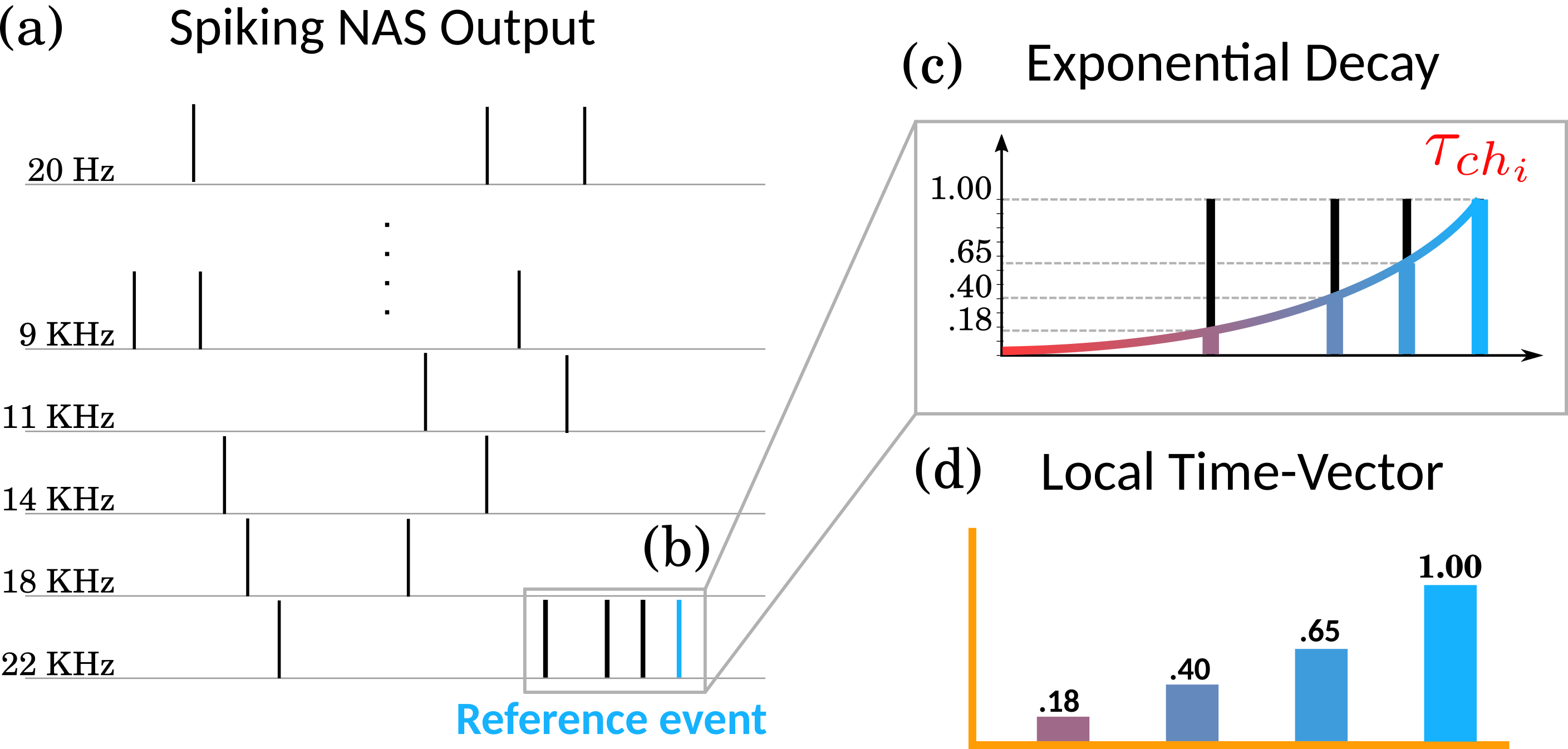}
    \caption{Example of a Local Time-Vector generation with n=4. Every time a new spike is produced by the NAS (a) we select the n-1 last events produced by the same channel $ch_i$ of the reference event (b). We interpolate an exponential decay with a time constant $\tau_{ch_i}$ (c) selected for the channel $ch_i$ and we define the Local Time vector as a vector containing the values obtained from (c). }
    \label{fig:Local Time-Vectors}
\end{figure}

A time vector is a descriptor similar to Time-Surfaces \cite{lagorce2016hots}, in which the temporal neighborhood of a certain event, is interpolated with an exponential decay.
Similarly to synaptic computation, this representation is useful to extract the temporal relations between subsequent events, and characterize patterns of activity by their relative spike timings.
In neuromorphic computation an event is a precisely timed digital signal, carrying temporal information along other information, like pixel coordinates (as for artificial retinas like ATIS \cite{posch2008asynchronous} or DAVIS \cite{brandli2014240}) channel index (as for the NAS \cite{jimenez2016binaural}).
Therefore, events produced by a NAS can be described by the tuple: 
\begin{equation}
ev_i = [t_i, ch_i] \qquad i \in \mathbb{N}
\label{eq:event}
\end{equation}
Where $t_i$ is the timestamp of a single $i$ event and $ch_i$ represents the index of the frequency tuned channel producing it. ($ch_i \in [0,31] \subset \mathbb{N}$) 
From it, we can define an ordered set $T_{ch_i}$ of timestamps occurring before the event $ev_i$ generating from the same channel:
\begin{equation}
T_{ch_i} = \{t_j|t_j<=t_i, t_{j-1}<t_j, ch_j=ch_i\}
\label{eq:channel_context}
\end{equation}
and then we can select the last n timestamps that we will use to generate a Local Time Vector of length n,
\begin{equation}
TL_i = \{t_k| t_k\in{T_{ch_i}}, k=\{j,j-1,j-2...j-n-1\}\}
\label{eq:local_context}
\end{equation}
so that applying an exponential decay with a time coefficient $\tau_{ch_i}$,
\begin{equation}
SL_i=e^{\frac{-(t_i-TL_i)}{\tau_{ch_i}}} 
\label{eq:local_vector}
\end{equation}
we create a Local Time Vector $SL_i$ for the reference event $ev_i$
Where $\tau_{ch_i}$ is the time coefficient selected for that channel,
and describes the temporal precision or sensitivity to the patterns produced by the NAS.
%$n$, the length of each time surface, it is chosen to be high enough to fit the envelope of the exponential decay under mean firing rate over the entire dataset for the most active channel.
The process to generate the Local time surfaces is summarized by the Figure \ref{fig:Local Time-Vectors}, which shows the process of generating a Time-Vector with $n=4$ spikes.

\subsubsection{Temporal coefficients:}
The NAS spiking output presents a difference in spiking rate, where lower frequency channels produce less spikes than higher ones, for the same sound magnitude or sound pressure. This relation is problematic for neuromorphic algorithms, as information is expected to be encoded in the relative timings of spikes. In the specific case of Local Time Vectors, the same pattern or sound pressure, will produce different Time-Vectors depending on the frequency of the channel, artificially increasing the complexity of the auditory feature encoding.

\begin{figure}[ht]
    \centering
    \includegraphics[scale=0.5]{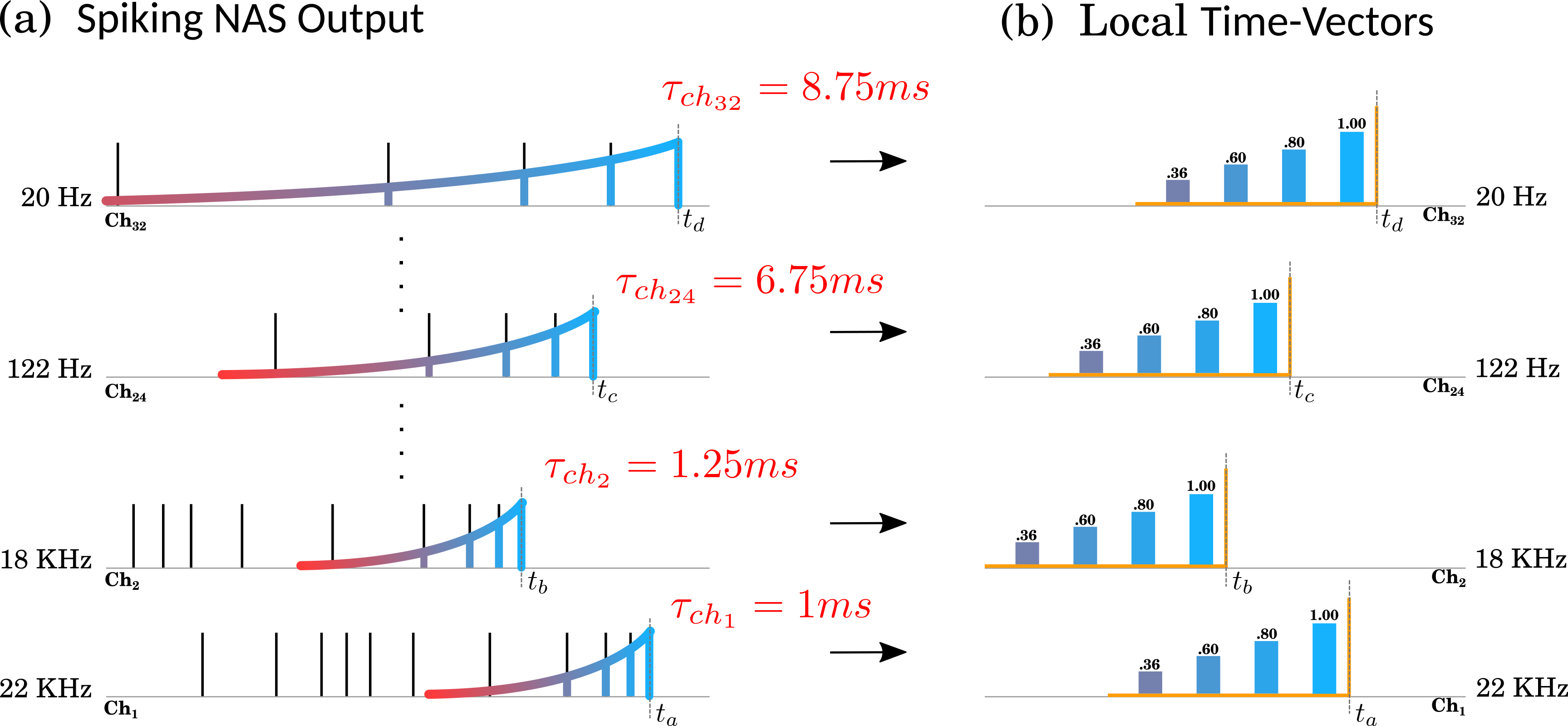}
    \caption{This figure shows the temporal scaling used to normalize the spike information coming from different channels. As different channels have different spiking rate for the same frequency amplitude (a), we select individual $\tau_{ch_i}$ so that the corresponding Local-Time vector will be the same. }
    \label{fig:Time_coeff}
\end{figure}

For this reason, when we create Local Time-Vectors for the Local Features Layer, we define a different $\tau_{ch_i}$ for each individual channels, so that the Time-Vectors will be normalized across different frequencies (Fig. \ref{fig:Time_coeff}) 
Because $\tau_{ch_i}$ is dependent on the channel index of the NAS but also serves as a fundamental parameter for the Local Features Layer we break it down in two terms:
\begin{equation}
\tau_{ch_i}=m_{ch_i} *\tau_{local}\\
\end{equation}
Where $\tau_{local}$ is the parameter used to control the overall temporal precision of the first layer, and $m_{ch_i}$ is a multiplicative factor that results from linear interpolation between the highest spiking frequency of channel 0 ($9\cdot10^{4}$ spikes/s) and the lowest frequency channel 32 ($2\cdot10^{4}$ spikes/s) \cite{jimenez2016binaural}.
\begin{equation}
m_{ch_i} = 2 + ch_i*(7/31)
\end{equation}

{}

\subsubsection{Learning and inferring local features}

\begin{figure}[ht]
    \centering
    \includegraphics[scale=0.60]{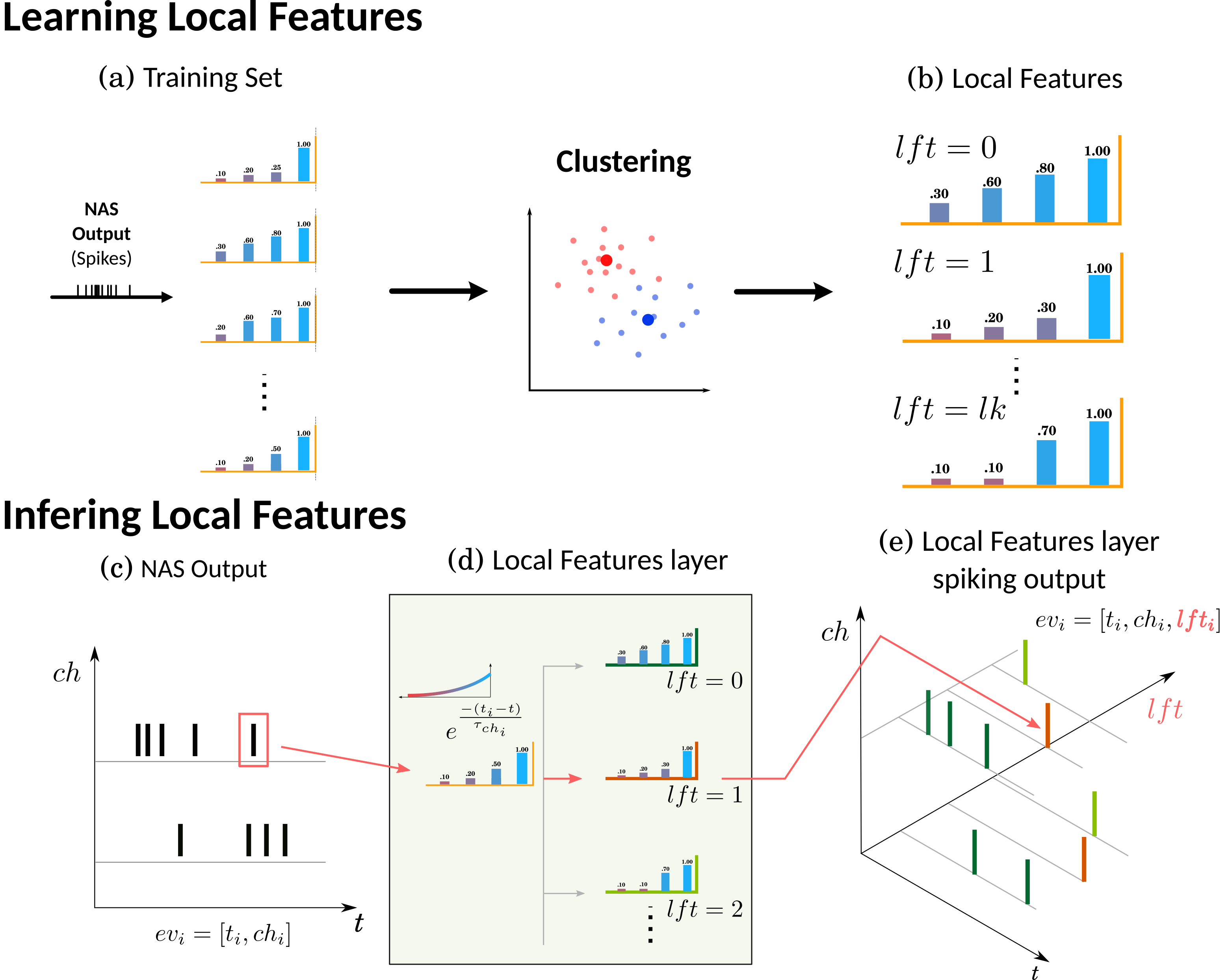}
    \caption{In this figure we show the two modalities in which the Local Feature Layer can operate (Learning and Infering). During learning, the layer clusters a Training set of Local Time-Vectors (a) and extract a set of lk features, were lk is the selected number of clusters for the local layer. During Infering, any incoming event (c) is classified by the closest cluster (d), and its class index represented by a new descriptor $lft_i$ (e).}
    \label{fig:Local_layer}
\end{figure}

In order to learn a useful representation of the Local Time-Vector generated by the NAS, we then cluster a training set as shown in Figure \ref{fig:Local_layer}a. To perform the clustering, we decided to use a faster and less memory intensive K-Means implementation of the Scikit-Learn python library \cite{scikit-learn}: Mini Batch K-means \cite{minibatchkmeans}. Although, other clustering algorithms are compatible with this type of network \cite{lagorce2016hots}, \cite{hots2018sparse}.
The centroids of the clustering algorithm are then considered features representing the multitude of different Local Time-Vectors generated by the NAS (Fig. \ref{fig:Local_layer}c). Every time that a new event is generated (Fig. \ref{fig:Local_layer}c), a Local Time-Vector is built and assigned to the closest center (Fig. \ref{fig:Local_layer}d); the result of this assignment is a new event with the same $t_i$ and $ch_i$ of the input one, but with an added descriptor called local feature $lft_i$ (Fig. \ref{fig:Local_layer}e), which stores the center index of the closest cluster:
\begin{equation}
ev_i = [t_i, ch_i, lft_i] \qquad i \in \mathbb{N}
\label{eq:local_event}
\end{equation}
Where $lft_i \in [0,lk] \subset \mathbb{N}$, and $lk$ is the selected number of clusters for the local layer.

{}

\subsection{Cross features layer} 
The second layer is used to extract temporal patterns of local features across channels. The main idea is that once we characterized the temporal evolution of single channels we are now ready to pool information from multiple frequencies. Since Local-Time vectors were extracting patterns from single channels, we now need to define a new descriptor that we call Cross Time-Vector, to identify patterns across channels. Similarly as we did with Local Time-vectors, we will then show how we learn and infer Cross Time-vectors in this layer.

\subsubsection{Cross Time-Vectors:} 
\begin{figure}[ht]
    \centering
    \includegraphics[scale=0.58]{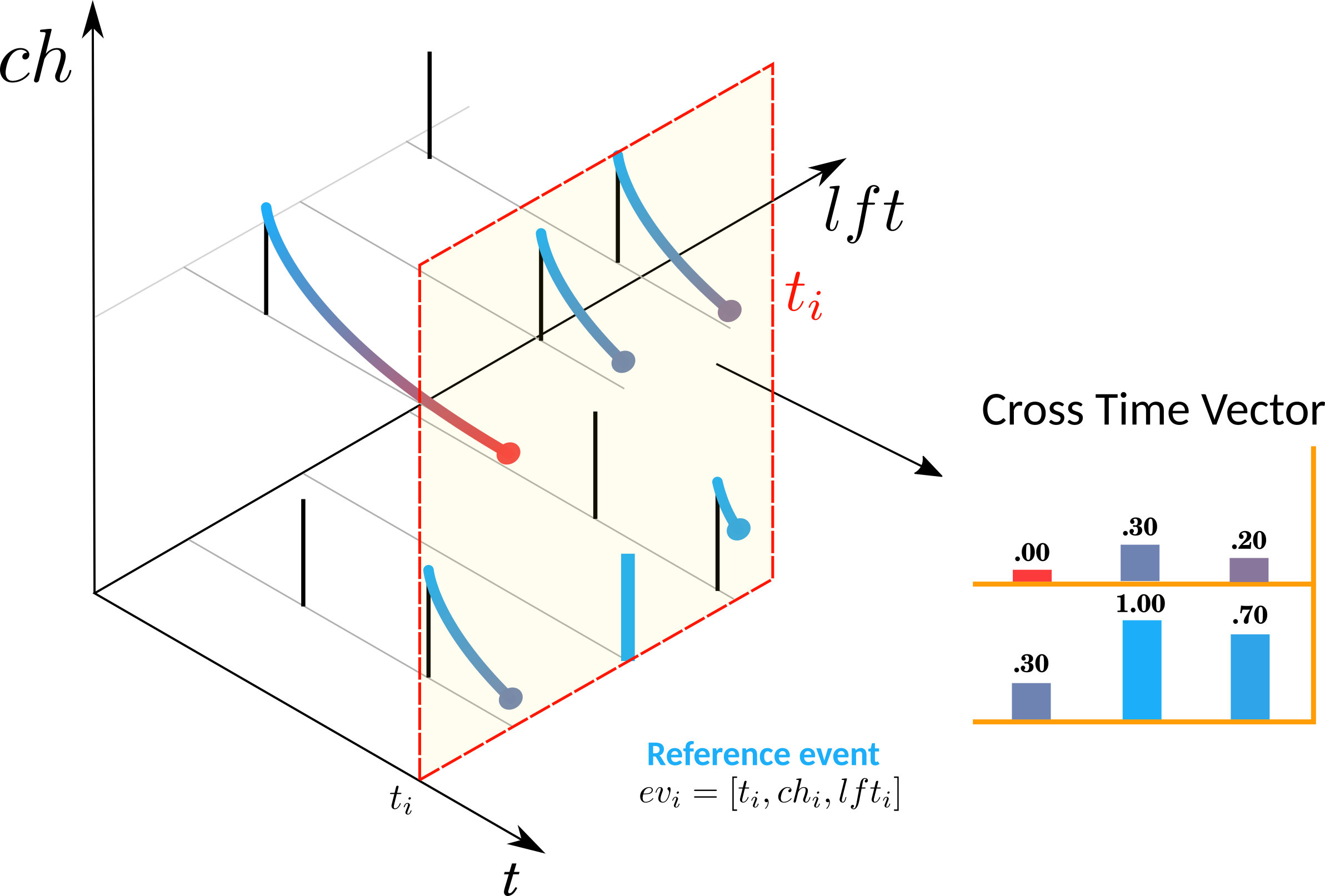}
    \caption{Example of Cross Time-Vectors generation. Each time the second layer receive a new event, the algorithm pools the last event of each channel and local feature. It then calculates the exponential decay for each event at the reference time $t_i$ (yellow surface) and then it composes a Cross Time-Vector containing all the decay values for all the combinations of channel/local feature.}
    \label{fig:Cross_generation}
\end{figure}

The Time-Surfaces used here have a different dimensionality than the Time-Vector presented previously, as events are now characterized by channel index $ch_i$ and local feature $lft_i$.
For this reason, we now define the temporal neighborhood of each new event $ev_i$ by this equation: 
\begin{equation}
TC_i = \max_{j \leq i}\{t_j \forall (n,m) : ch_j=n, lft_j=m \}
\end{equation}
Composed by all the last events before $t_i$ produced by every channel and local features.
So that the resulting Cross Time-Surfaces is:
\begin{equation}
SC_i=e^{\frac{-(t_i-TC_i)}{\tau_{cr}}} 
\end{equation}
Where $\tau_{cr}$ is the temporal coefficient used to tune Cross Time-Surfaces to extract information at the temporal scale of interest.
A graphical depiction of this process is summarised in Figure \ref{fig:Cross_generation}.
{}

\subsubsection{Learning and inferring cross features}
Similarly for the Local features layer we use the same Kmeans algorithm \cite{minibatchkmeans} to learn a set of clusters of length from a training set of cross time-vectors. When inferring, every incoming event will generate a cross time-vector and will be assigned to the closest cluster generating a new event:
\begin{equation}
ev_i = [t_i, cft_i] \qquad i \in \mathbb{N}
\label{eq:cross_event}
\end{equation}
Where $cft_i \in [0,ck] \subset \mathbb{N}$ is the index of the matching cross-feature, and $ck$ is the total number of clusters for the cross layer.

\subsection{Activity Histogram} 
\begin{figure}[ht]
    \centering
    \includegraphics[scale=0.42]{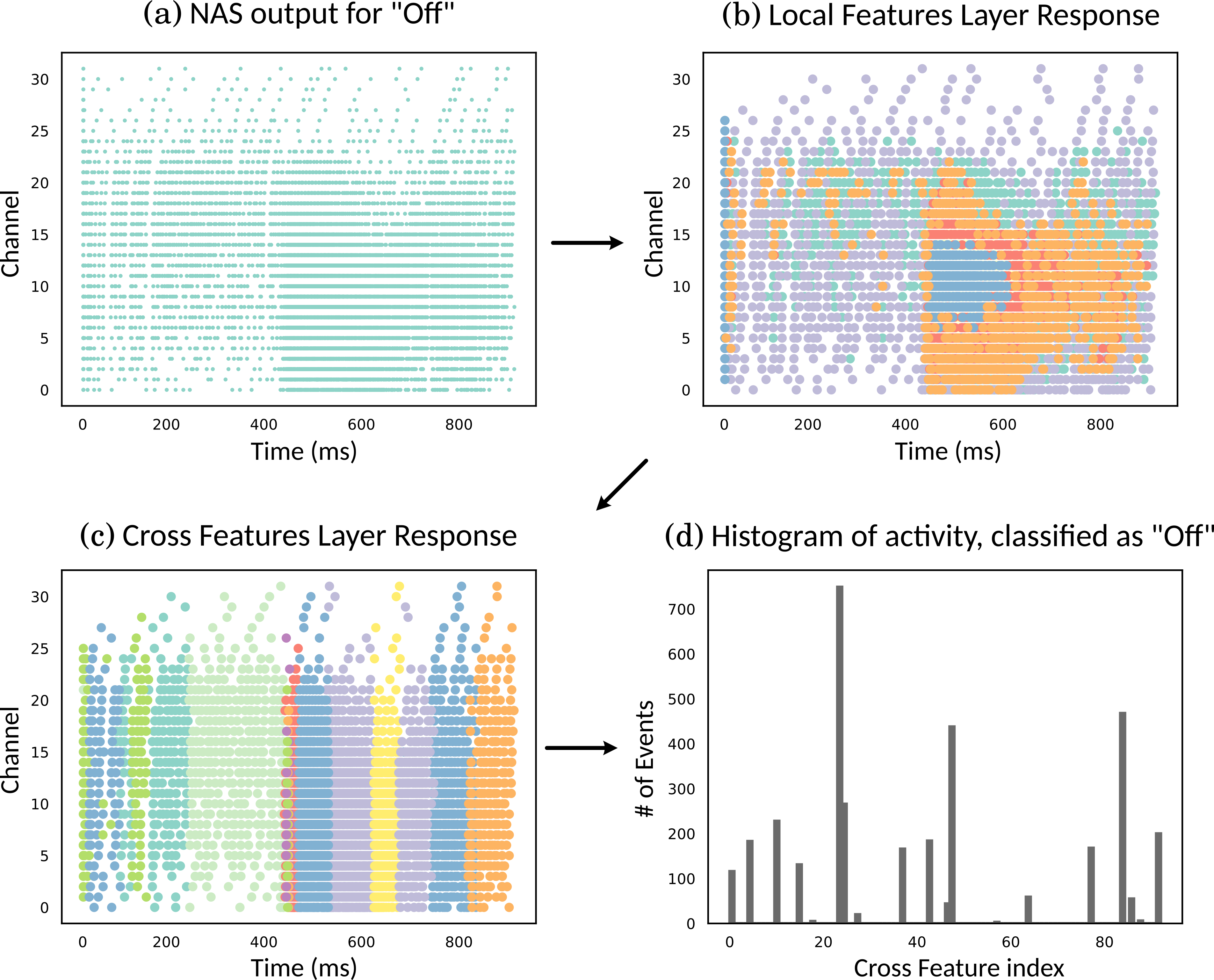}
    \caption{The response of the network to a recording of the spoken word ''Off''. The NAS spiking activity (a) is classified by the Local Feature Layer which response has been color-coded by the index of the corresponding local feature (b). Each event is then classified by Cross Feature by the Cross Feature layer (c), and finally the network response is collected by the Histogram of activity that shows the total number of events per Cross Feature.}
    \label{fig:Net_response}
\end{figure}

One way to classify the output of event based networks is to build histograms of activity \cite{lagorce2016hots}. For a single audio recording or time window, we can build an histogram $\mathcal{H}$ by counting the number of events generated by the network per cross feature $cft$. We can then apply different classifiers on activity histograms in order to recognize spoken words, phonemes or sounds.
In Figure \ref{fig:Net_response} we show the network response to a spoken word ("off"). 

{}

\subsection{Classifier} 
In this paper we test the network with three different classifiers: Euclidean distance, Normalised Euclidean distance, and a Multi-Layer-Perceptron (MLP) model. In the first case, we compute the output histograms of the training set of the dataset, and then we average them by class, creating a set of ''prototypes'' $\mathcal{R}$. Then for every new recording we can generate a new histogram and calculate its euclidean distance to the prototype:
\begin{equation}
d(\mathcal{H}_{recording},\mathcal{R}_{class}) = \| \mathcal{H}_{recording}-\mathcal{R}_{class} \|
\label{eq:eucl_class}
\end{equation}
We then select the class with the minor distance to the recording.
In the case of Normalized Euclidean distance we divide the histogram of each recording by the total number of events generated, also creating a new set of normalized prototypes $\mathcal{NR}$:
\begin{equation}
dn(\mathcal{H}_{recording},\mathcal{NR}_{class}) = \| \frac{\mathcal{H}_{recording}}{card(\mathcal{H}_{recording})}-\mathcal{NR}_{class} \|
\label{eq:norm_eucl_class}
\end{equation}
where card indicates the number of events of the recording.
Finally we also use an MLP classifier to classify the histograms of training and test sets.

\section{Results}
We test the network on the yes/no words we recorded from the Google's Speech Command dataset \cite{warden2018speech}. We randomly select 3740 ''yes'' recordings and 3740 ''no'' recordings. We then use a random sample of 70$\%$ for training set and 30$\%$  as testing set, while keeping the same number of samples for both classes.
After testing different combinations of parameters we set $t_{local}=1ms$, $n=5$, and $lk=6$ for the Local features layer, and $t_{cr}=200ms$, and $ck=96$ for the Cross features layer. Finally for the MLP classifier we select a network with a single hidden layer of 30 ''RelU'' neurons. The result of this setup are reported in Table~\ref{tab:yes_no_Results}.

\begin{table}[]
    \centering
    \begin{tabular}{c|c|c}
      \textbf{Classifiers}         & Train set accuracy & Test set accuracy \\
      Eucl Histogram               & 74.66\%            &  74.42\% \\
      Norm Eucl Histogram          & 75.06\%            &  73.80\% \\
      MLP                          & 83.41\%            &  82.55\%
    \end{tabular}
    \caption{Results of the ''yes''/''no'' classification test.}
    \label{tab:yes_no_Results}
\end{table}

\section{Conclusions}
We presented a neuromorphic network capable of unsupervised auditory features extraction, compatible with a digital neuromorphic cochlea \cite{jimenez2016binaural}. We then coupled the network with three classifiers and tested it on a subset of the Google's Speech Commands dataset. The preliminary results showed that network was able to recognize simple commands for the Google's Speech Commands dataset with good accuracy, even when paired with weak classifiers, suggesting that the set of Local and Cross features extracted were informative for the task of word recognition. In order to completely evaluate the network we still need to expand the result section with the rest of Google's Speech Commands dataset and test the network with other datasets as well. Nevertheless, the compactness of the architecture, the small number of features per layer ($lk=6$, $ck=96$) and the simplicity of the classifiers make this network a good candidate for edge-computing sound recognition/NLP devices.

%Apple for its Siri ''wakeword'' in 2017 on used a far more complex 5 layers recurrent neural network constantly running on a always-on chip.

\bibliographystyle{ieeetr}
\bibliography{biblio}

\begin{thebibliography}{10}

\bibitem{edgeAI}
M.~G.~S. Murshed, C.~Murphy, D.~Hou, N.~Khan, G.~Ananthanarayanan, and
  F.~Hussain, ``Machine learning at the network edge: A survey,'' {\em ACM
  Comput. Surv.}, vol.~54, oct 2021.

\bibitem{IOT}
M.~S. Murshed, C.~Murphy, D.~Hou, N.~Khan, G.~Ananthanarayanan, and F.~Hussain,
  ``Machine learning at the network edge: A survey,'' {\em ACM Computing
  Surveys (CSUR)}, vol.~54, no.~8, pp.~1--37, 2021.

\bibitem{market2018global}
A.~Market, ``Global industry analysis, size, share, growth, trends, and
  forecast 2017--2025,'' {\em URL: https://www. transparencymarketresearch.
  com/logistics-market. html (Last accessed: 16.01. 2018)}, 2018.

\bibitem{medicalkumari2017increasing}
P.~Kumari, L.~Mathew, and P.~Syal, ``Increasing trend of wearables and
  multimodal interface for human activity monitoring: A review,'' {\em
  Biosensors and Bioelectronics}, vol.~90, pp.~298--307, 2017.

\bibitem{harrop2016mobile}
J.~Harrop, ``Mobile robotics expected to fuel industry growth in robotics
  market,'' {\em Control Engineering}, vol.~62, no.~2, pp.~20--21, 2016.

\bibitem{chen2019deep}
J.~Chen and X.~Ran, ``Deep learning with edge computing: A review.,'' {\em
  Proc. IEEE}, vol.~107, no.~8, pp.~1655--1674, 2019.

\bibitem{mead1990neuromorphic}
C.~Mead, ``Neuromorphic electronic systems,'' {\em Proceedings of the IEEE},
  vol.~78, no.~10, pp.~1629--1636, 1990.

\bibitem{indiveri2011neuromorphic}
G.~Indiveri, B.~Linares-Barranco, T.~J. Hamilton, A.~Van~Schaik,
  R.~Etienne-Cummings, T.~Delbruck, S.-C. Liu, P.~Dudek, P.~H{\"a}fliger,
  S.~Renaud, {\em et~al.}, ``Neuromorphic silicon neuron circuits,'' {\em
  Frontiers in neuroscience}, vol.~5, p.~73, 2011.

\bibitem{NeuromorphNature}
K.~Roy, A.~Jaiswal, and P.~Panda, ``Towards spike-based machine intelligence
  with neuromorphic computing,'' {\em Nature}, vol.~575, no.~7784,
  pp.~607--617, 2019.

\bibitem{zhu2020comprehensiveReview}
J.~Zhu, T.~Zhang, Y.~Yang, and R.~Huang, ``A comprehensive review on emerging
  artificial neuromorphic devices,'' {\em Applied Physics Reviews}, vol.~7,
  no.~1, p.~011312, 2020.

\bibitem{yang20160}
M.~Yang, C.-H. Chien, T.~Delbruck, and S.-C. Liu, ``A 0.5 {V} 55$\mu${W}
  64$\times$2 channel binaural silicon cochlea for event-driven stereo-audio
  sensing,'' {\em IEEE Journal of Solid-State Circuits}, vol.~51, no.~11,
  pp.~2554--2569, 2016.

\bibitem{chan2007aer}
V.~Chan, S.-C. Liu, and A.~van Schaik, ``Aer ear: A matched silicon cochlea
  pair with address event representation interface,'' {\em IEEE Transactions on
  Circuits and Systems I: Regular Papers}, vol.~54, no.~1, pp.~48--59, 2007.

\bibitem{xu2018fpga}
Y.~Xu, C.~S. Thakur, R.~K. Singh, T.~J. Hamilton, R.~M. Wang, and A.~van
  Schaik, ``A fpga implementation of the car-fac cochlear model,'' {\em
  Frontiers in neuroscience}, vol.~12, p.~198, 2018.

\bibitem{jimenez2016binaural}
A.~Jim{\'e}nez-Fern{\'a}ndez, E.~Cerezuela-Escudero, L.~Mir{\'o}-Amarante,
  M.~J. Dom{\'\i}nguez-Morales, F.~de~As{\'\i}s G{\'o}mez-Rodr{\'\i}guez,
  A.~Linares-Barranco, and G.~Jim{\'e}nez-Moreno, ``A binaural neuromorphic
  auditory sensor for {FPGA}: A spike signal processing approach,'' {\em IEEE
  transactions on neural networks and learning systems}, vol.~28, no.~4,
  pp.~804--818, 2016.

\bibitem{o2013real}
P.~O'Connor, D.~Neil, S.-C. Liu, T.~Delbruck, and M.~Pfeiffer, ``Real-time
  classification and sensor fusion with a spiking deep belief network,'' {\em
  Frontiers in neuroscience}, vol.~7, p.~178, 2013.

\bibitem{schoepe2019neuromorphic}
T.~Schoepe, D.~Gutierrez-Galan, J.~P. Dominguez-Morales, A.~Jimenez-Fernandez,
  A.~Linares-Barranco, and E.~Chicca, ``Neuromorphic sensory integration for
  combining sound source localization and collision avoidance,'' in {\em 2019
  IEEE Biomedical Circuits and Systems Conference (BioCAS)}, pp.~1--4, IEEE,
  2019.

\bibitem{lagorce2016hots}
X.~Lagorce, G.~Orchard, F.~Galluppi, B.~E. Shi, and R.~B. Benosman, ``Hots: a
  hierarchy of event-based time-surfaces for pattern recognition,'' {\em IEEE
  transactions on pattern analysis and machine intelligence}, vol.~39, no.~7,
  pp.~1346--1359, 2016.

\bibitem{hots2018sparse}
G.~Haessig and R.~Benosman, ``A sparse coding multi-scale precise-timing
  machine learning algorithm for neuromorphic event-based sensors,'' in {\em
  Micro-and Nanotechnology Sensors, Systems, and Applications X}, vol.~10639,
  p.~106391U, International Society for Optics and Photonics, 2018.

\bibitem{warden2018speech}
P.~Warden, ``Speech commands: A dataset for limited-vocabulary speech
  recognition,'' {\em arXiv preprint arXiv:1804.03209}, 2018.

\bibitem{lyon1982computational}
R.~Lyon, ``A computational model of filtering, detection, and compression in
  the cochlea,'' in {\em ICASSP'82. IEEE International Conference on Acoustics,
  Speech, and Signal Processing}, vol.~7, pp.~1282--1285, IEEE, 1982.

\bibitem{jimenez2010building}
A.~Jimenez-Fernandez, A.~Linares-Barranco, R.~Paz-Vicente, G.~Jim{\'e}nez, and
  A.~Civit, ``Building blocks for spikes signals processing,'' in {\em The 2010
  International Joint Conference on Neural Networks (IJCNN)}, pp.~1--8, IEEE,
  2010.

\bibitem{jimenez2012neuro}
A.~Jimenez-Fernandez, G.~Jimenez-Moreno, A.~Linares-Barranco, M.~J.
  Dominguez-Morales, R.~Paz-Vicente, and A.~Civit-Balcells, ``A neuro-inspired
  spike-based pid motor controller for multi-motor robots with low cost
  fpgas,'' {\em Sensors}, vol.~12, no.~4, pp.~3831--3856, 2012.

\bibitem{AER}
``The {A}dress-{E}vent {R}epresentation communication protocol.''

\bibitem{paz2009synthetic}
R.~Paz-Vicente, A.~Linares-Barranco, A.~Jimenez-Fernandez, G.~Jimenez-Moreno,
  and A.~Civit-Balcells, ``{Synthetic retina for AER systems development},'' in
  {\em 2009 IEEE/ACS International Conference on Computer Systems and
  Applications}, pp.~907--912, IEEE, 2009.

\bibitem{furber2014spinnaker}
S.~B. Furber, F.~Galluppi, S.~Temple, and L.~A. Plana, ``The spinnaker
  project,'' {\em Proceedings of the IEEE}, vol.~102, no.~5, pp.~652--665,
  2014.

\bibitem{rosen2011signals}
S.~Rosen and P.~Howell, {\em Signals and systems for speech and hearing},
  vol.~29.
\newblock Brill, 2011.

\bibitem{berner20075}
R.~Berner, T.~Delbruck, A.~Civit-Balcells, and A.~Linares-Barranco, ``{A 5 Meps
  \$100 USB2.0 address-event monitor-sequencer interface},'' in {\em 2007 IEEE
  International Symposium on Circuits and Systems}, pp.~2451--2454, IEEE, 2007.

\bibitem{posch2008asynchronous}
C.~Posch, D.~Matolin, and R.~Wohlgenannt, ``An asynchronous time-based image
  sensor,'' in {\em 2008 IEEE International Symposium on Circuits and Systems},
  pp.~2130--2133, IEEE, 2008.

\bibitem{brandli2014240}
C.~Brandli, R.~Berner, M.~Yang, S.-C. Liu, and T.~Delbruck, ``A 240$\times$ 180
  130 db 3 $\mu$s latency global shutter spatiotemporal vision sensor,'' {\em
  IEEE Journal of Solid-State Circuits}, vol.~49, no.~10, pp.~2333--2341, 2014.

\bibitem{scikit-learn}
F.~Pedregosa, G.~Varoquaux, A.~Gramfort, V.~Michel, B.~Thirion, O.~Grisel,
  M.~Blondel, P.~Prettenhofer, R.~Weiss, V.~Dubourg, J.~Vanderplas, A.~Passos,
  D.~Cournapeau, M.~Brucher, M.~Perrot, and E.~Duchesnay, ``Scikit-learn:
  Machine learning in {P}ython,'' {\em Journal of Machine Learning Research},
  vol.~12, pp.~2825--2830, 2011.

\bibitem{minibatchkmeans}
D.~Sculley, ``Web-scale k-means clustering,'' in {\em Proceedings of the 19th
  international conference on World wide web}, pp.~1177--1178, 2010.

\end{thebibliography}
\end{document}